\documentclass[letterpaper]{article} 
\usepackage{aaai2026}  
\usepackage{times}  
\usepackage{amsfonts}
\usepackage{helvet}  
\usepackage{courier}  
\usepackage[hyphens]{url}  
\usepackage{graphicx} 
\usepackage{bm} 
\usepackage{siunitx}
\usepackage{natbib}  
\usepackage{caption} 
\usepackage{algorithm}
\usepackage{algorithmic}
\usepackage{subcaption}
\usepackage{amsmath}
\usepackage{multirow}      
\usepackage{booktabs}      
\usepackage{amsmath}       
\usepackage{makecell}  
\usepackage{listings}
\DeclareCaptionStyle{ruled}{labelfont=normalfont,labelsep=colon,strut=off} 
\floatstyle{ruled}
\newfloat{listing}{tb}{lst}{}
\floatname{listing}{Listing}
\title{GloCTM: Cross-Lingual Topic Modeling via a Global Context Space}
\author{
    Nguyen Tien Phat\textsuperscript{\rm 1}\equalcontrib,
    Ngo Vu Minh\textsuperscript{\rm 1}\equalcontrib,
    Linh Van Ngo\textsuperscript{\rm 1}\thanks{Corresponding Author.},
    Nguyen Thi Ngoc Diep\textsuperscript{\rm 2},\\
    Thien Huu Nguyen\textsuperscript{\rm 3}
}

\affiliations{
    \textsuperscript{\rm 1} Hanoi University of Science and Technology, Vietnam\\
    \textsuperscript{\rm 2} VNU University of Engineering and Technology, Vietnam\\
    \textsuperscript{\rm 3} University of Oregon, US\\
    \{phat.nt235544, minh.nv230084\}@sis.hust.edu.vn,\;
    linhnv@soict.edu.vn,\;
    ngocdiep@vnu.edu.vn,\;
    thienn@uoregon.edu
}

\begin{document}
\maketitle

\begin{abstract}
Cross-lingual topic modeling seeks to uncover coherent and semantically aligned topics across languages—a task central to multilingual understanding. Yet most existing models learn topics in disjoint, language-specific spaces and rely on alignment mechanisms (e.g., bilingual dictionaries) that often fail to capture deep cross-lingual semantics, resulting in loosely connected topic spaces. Moreover, these approaches often overlook the rich semantic signals embedded in multilingual pretrained representations, further limiting their ability to capture fine-grained alignment. We introduce \textbf{GloCTM} (\textbf{Glo}bal Context Space for \textbf{C}ross-Lingual \textbf{T}opic \textbf{M}odel), a novel framework that enforces cross-lingual topic alignment through a unified semantic space spanning the entire model pipeline. GloCTM constructs enriched input representations by expanding bag-of-words with cross-lingual lexical neighborhoods, and infers topic proportions using both local and global encoders, with their latent representations aligned through internal regularization. At the output level, the global topic-word distribution, defined over the combined vocabulary, structurally synchronizes topic meanings across languages. To further ground topics in deep semantic space, GloCTM incorporates a Centered Kernel Alignment (CKA) loss that aligns the latent topic space with multilingual contextual embeddings. Experiments across multiple benchmarks demonstrate that GloCTM significantly improves topic coherence and cross-lingual alignment, outperforming strong baselines. 


\end{abstract}
\begin{links}
    \link{Code}{https://github.com/tienphat140205/GloCTM}

\end{links}


\section{Introduction}

\begin{figure}
    \centering
    \includegraphics[width=0.9\linewidth]{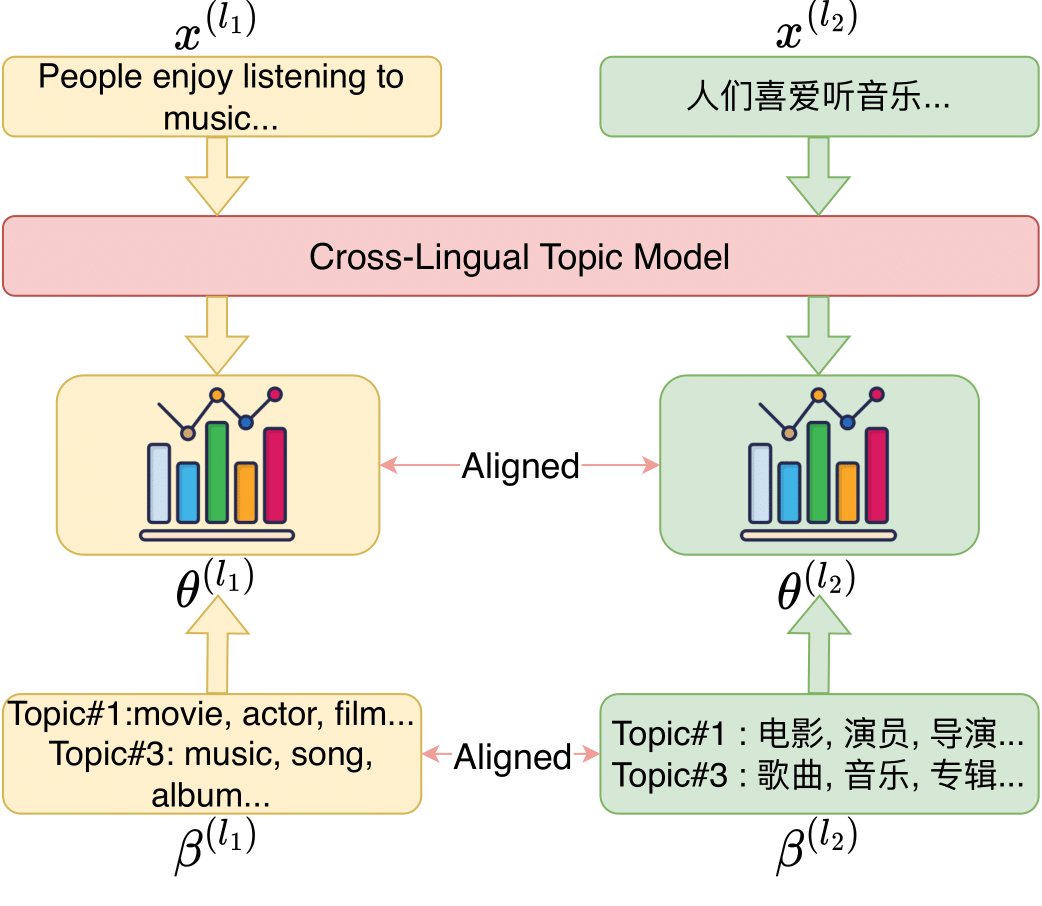}
    \caption{Conceptual overview of Cross-Lingual Topic Modeling (CLTM)}
    \label{fig:cross-lingual}
\end{figure}

Uncovering latent thematic structures from large-scale text corpora is a foundational task in natural language processing, with topic modeling (TM) serving as a powerful technique for distilling meaning from data \cite{blei2003lda}. The challenge intensifies in a globalized world, where understanding shared narratives across different languages is paramount. Cross-Lingual Topic Modeling (CLTM) addresses this challenge, as illustrated in Figure \ref{fig:cross-lingual}. The objective is to discover coherent topics that maintain their semantic identity across linguistic boundaries \cite{MimnoWNSM09,boyd2009,Ni2009}. For instance, given documents sharing the same semantic content across different languages, an ideal CLTM infers similar document-topic proportions ($\theta$). This requires the underlying topics themselves, defined by their word distributions ($\beta$), to be robustly aligned, such that a single topic index corresponds to the same concept in each language. This capability thereby enables large-scale cross-cultural analysis and global knowledge discovery.

\begin{table}
    \centering
    \includegraphics[width=1\linewidth]{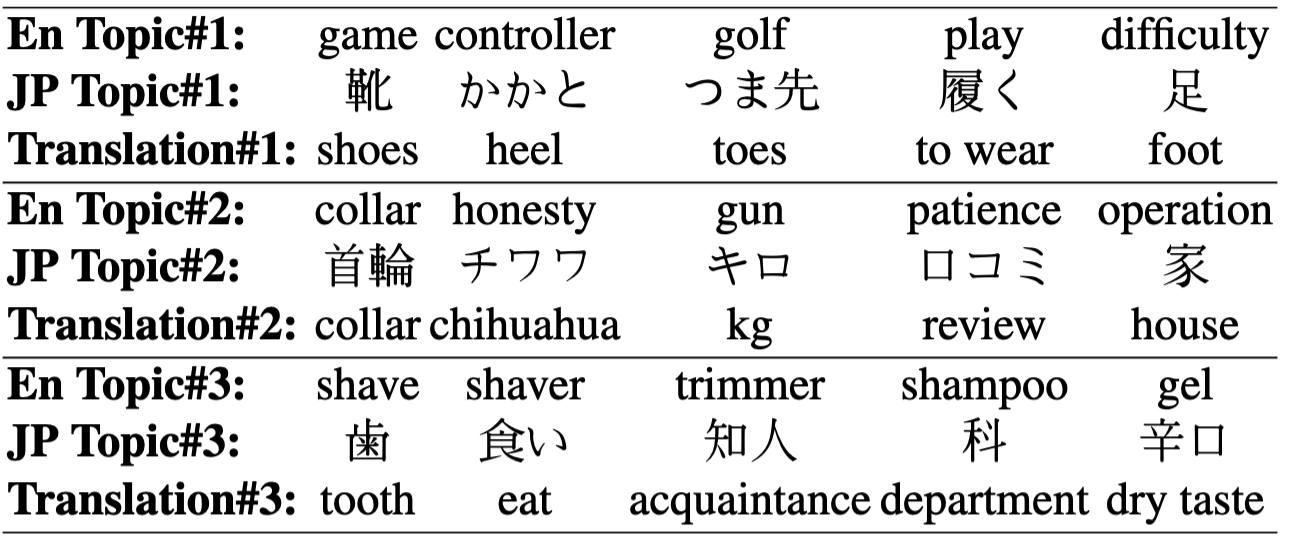}
    \caption{Examples of misaligned topics generated by InfoCTM \cite{infoctm} across English (En) and Japanese (JP). The words grouped under each topic index show a clear semantic drift between the two languages.}
    \label{fig:miss-aligned}
\end{table}
Early efforts in cross-lingual topic modeling (CLTM) often relied on parallel corpora to learn aligned topics across languages \cite{MimnoWNSM09}. However, such data is scarce and expensive to obtain, particularly for low-resource language pairs. As a result, most subsequent approaches turn to bilingual dictionaries as a more practical alternative \cite{JagarlamudiD10,Hu2014,DBLP:conf/acl/ShiLBX16, YuanDY18, WuLZM20, infoctm}. While effective to some extent, this strategy introduces two key challenges. First, dictionary coverage is often limited \cite{ChangH21}, leading to poor topic alignment \cite{JagarlamudiD10,HaoP18}. Second, according to the analysis in \cite{infoctm}, the direct alignment of topic-word distributions in models such as NMTM \cite{WuLZM20} and MCTA \cite{DBLP:conf/acl/ShiLBX16} can lead to a trivial solution. In this scenario, topic representations degenerate, causing the resulting topics to collapse and lose their interpretability. While recent models such as InfoCTM \cite{infoctm} have partially alleviated topic collapse and low dictionary coverage, their underlying architectural philosophy still inherits fundamental limitations from the broader CLTM landscape. A key concern is that these models continue to learn topic proportions ($\theta$) and topic-word distributions ($\beta$) in separate, disjoint spaces for each language, relying on an auxiliary loss function to indirectly bridge the gap between these otherwise disconnected language spaces. This architectural separation is the root cause of why topic alignment remains insufficiently robust. As illustrated in Figure \ref{fig:miss-aligned}, this frequently leads to a "semantic drift," where a single topic index can correspond to a coherent theme in one language (e.g., "video games" in English) but a completely unrelated one in another (e.g., "footwear" in Japanese). This highlights the need for a framework that enforces alignment structurally, rather than handling it as an external or secondary component.

Moreover, many models still depend on static dictionaries with limited coverage and shallow lexical cues, while overlooking the rich semantic signals offered by multilingual pretrained language models. Recent works \cite{BianchiTHNF21, MuellerD21} utilize multilingual BERT for topic modeling over multilingual corpora, but their goal is not to learn aligned cross-lingual topics, and thus they fall outside the scope of cross-lingual topic modeling. Another line of work, the clustering-based refinement method by \citep{refining}, falls short, as it discovers latent groupings from multilingual embeddings but fails to model topic distributions, incorporate a generative structure, or enforce cross-lingual alignment. These trends highlight a critical gap: the absence of a unified framework that integrates multilingual embeddings with interpretable, aligned, and generative topic modeling. 

To overcome these limitations, we introduce \textbf{GloCTM}, a dual-pathway neural topic model that brings topic alignment to the core of its design. Rather than relying on auxiliary losses to bridge disjoint language spaces, GloCTM constructs a shared semantic space from the outset. It begins by enriching each document with semantically related words from both languages (identified via multilingual pretrained embeddings) to form a dynamic global bag-of-words representation. This enriched input is passed through a global encoder, which infers a unified topic proportion vector ($\theta_{\text{global}}$), later decoded via a global topic-word matrix ($\beta_{\text{global}}$) that is constructed from the topic-word matrices of the individual language pathways to structurally enforce cross-lingual topic sharing. At the same time, language-specific local encoders capture monolingual subtleties, while a KL divergence loss encourages their outputs to stay aligned with the global topic view. This joint architecture allows GloCTM to balance language-specific expressiveness with semantic coherence across languages—offering a principled, end-to-end solution to cross-lingual topic modeling. 

In light of the rapid advances in multilingual pretrained language models, incorporating their semantic knowledge into cross-lingual topic modeling has become not only natural but essential. These embeddings provide a rich, language-agnostic semantic space that can inform topic inference across languages. To harness this, we introduce a final calibration mechanism based on Centered Kernel Alignment (CKA)~\cite{cortes2012algorithms, kornblith2019similarity}. Specifically, we apply a CKA loss to align the model’s latent topic proportion vectors with the deep semantic representations produced by multilingual pretrained language models (PLMs). This regularization enhances the semantic structure of the topic proportions (\(\theta\)), enabling them to reflect cross-lingual meaning beyond lexical overlap. As a result, \textbf{GloCTM} produces not only structurally aligned topics, but also semantically coherent ones as a natural consequence of improved latent representations.
Our main contributions are summarized as follows:
\begin{itemize}
    \item We introduce \textbf{GloCTM}, a novel dual-pathway framework for cross-lingual topic modeling that combines a global and local VAE architecture, while leveraging multilingual pretrained embeddings to guide topic learning across languages.
    
    \item We propose a \textbf{dynamic global bag-of-words construction} that enriches each document with semantically related words from both languages, enabling alignment-aware input representations grounded in lexical neighborhoods.
    
        \item We enforce robust alignment through a dual strategy: structurally, by decoding through a \textbf{unified topic-word matrix}, and representationally, through a \textbf{KL divergence loss} for internal consistency and a \textbf{Centered Kernel Alignment (CKA) loss} for semantic grounding.
    
    \item We empirically demonstrate that \textbf{GloCTM} consistently outperforms strong baselines on both topic diversity and cross-lingual alignment across multiple datasets and language pairs.
\end{itemize}
\begin{figure*}
    \centering
    \includegraphics[width=0.85\linewidth]{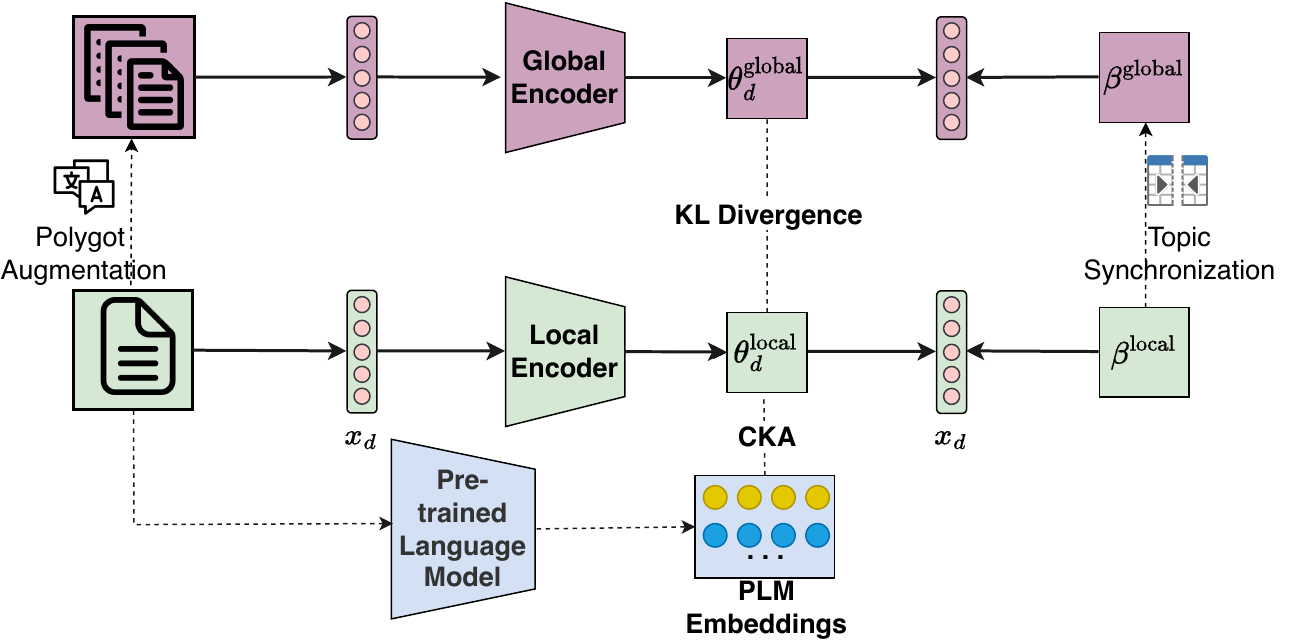}
\caption{The dual-pathway architecture of GloCTM. The model integrates a Local Pathway for language-specific features with a Global Pathway for unified topic learning. The Global Pathway is uniquely fed by a semantically-enriched input created via Polyglot Augmentation. Alignment is enforced through three core mechanisms: Topic Synchronization ($\beta$), KL Divergence ($\theta_{local} \leftrightarrow \theta_{global}$), and CKA loss with PLM embeddings.}

    \label{fig:model}
\end{figure*}
\section{Related Work}
\textbf{Cross-Lingual Topic Modeling.} Topic models aim to uncover latent semantic structures in large-scale corpora, with Latent Dirichlet Allocation (LDA) \cite{blei2003lda} as a foundational example. Neural topic models (NTMs) extend this framework using variational autoencoders (e.g., NVDM, ProdLDA \cite{srivastava2017prodlda}) and contextual embeddings \cite{dieng2020etm,BianchiTHNF21}.

Cross-lingual topic modeling (CLTM) adapts this idea to multilingual data, seeking coherent and aligned topics across languages. Early CLTM relied on parallel corpora \cite{MimnoWNSM09}, which are scarce. Dictionary-based methods became standard \cite{JagarlamudiD10, Hu2014, DBLP:conf/acl/ShiLBX16,YuanDY18,YangBR17,YangBR19,WuLZM20}, but suffer from limited coverage, weak semantics, and topic collapse. InfoCTM \cite{infoctm} improves topic diversity via contrastive learning and introduces a novel dictionary expansion technique, Cross-lingual Vocabulary Linking (CVL). Recently, the XTRA framework \cite{xtra} combines bag-of-words modelling with multilingual embeddings and dual contrastive alignment of document-topic and topic-word distributions to yield interpretable and well-aligned cross-lingual topics. Other work uses pretrained transformers for zero-shot modeling \cite{BianchiTHNF21,MuellerD21}, though lacking mechanisms for interpretable topic-word distributions. Refining Dimensions \cite{refining}, based on BERTopic \cite{grootendorst2022bertopic}, improves alignment by removing noisy embedding dimensions but does not infer topic or document distributions. These limitations highlight a key gap: current CLTM approaches often trade off interpretability or rely on shallow alignment, motivating the need for a principled, generative model that ensures both semantic depth and cross-lingual consistency.

\textbf{Centered Kernel Alignment (CKA).} CKA, introduced in kernel learning \cite{cortes2012algorithms}, extends the Hilbert-Schmidt Independence Criterion by computing normalized similarity between centered Gram matrices. Popularized in deep learning by Kornblith et al. \cite{kornblith2019similarity}, it captures representation similarity across layers and architectures. CKA has since been used in transfer learning \cite{bonheme2023vae, matsoukas2022medical}, architecture analysis \cite{raghu2021vision}, knowledge distillation \cite{saha2022cka, zhou2024rcka}, and model pruning \cite{pons2024pruning}. Despite its potential, CKA remains underexplored in topic modeling, where it could align document-topic and embedding-based representations across languages.

\section{Methodology}
\label{sec:methodology}
We propose \textbf{GloCTM}, a dual-pathway Variational Autoencoder (VAE) framework comprising parallel local and global branches. The local branch consists of two language-specific VAEs—one for each language—while the global branch is shared and receives inputs enriched via our \textbf{Polyglot Augmentation} method. As illustrated in Figure~\ref{fig:model}, the model’s robustness arises from three core mechanisms: \textbf{Topic Synchronization} through a unified decoder, a \textbf{KL Divergence} loss for internal consistency, and a \textbf{Centered Kernel Alignment (CKA)} loss for semantic grounding in pre-trained embeddings. The following sections detail these components.
\subsection{Notations}

We consider a multilingual corpus $\mathcal{C}$ with documents in two languages, $L_1$ and $L_2$, and aim to discover $K$ shared topics. Each document $d$ in language $L_l$ is represented by: (i) a BoW vector $\mathbf{x}_d^{(l)} \in \mathbb{R}^{|V^{(l)}|}$, (ii) a cross-lingually enriched vector $\mathbf{g}_d^{(l)} \in \mathbb{R}^{|V^{(1)}| + |V^{(2)}|}$, and (iii) an $M$-dimensional pretrained embedding $\mathbf{e}_d^{(l)}$. Parallel local and global encoders map these inputs to latent topic proportions, $\boldsymbol{\theta}_d^{(l,\text{local})}$ and $\boldsymbol{\theta}_d^{(l,\text{global})}$, which are decoded via language-specific matrices $\boldsymbol{\beta}^{(l)}$ and a unified matrix $\boldsymbol{\beta}^{(\text{global})}$ obtained by concatenating them.
\subsection{Polyglot Augmentation for a Pre-Aligned Global Context}

A key limitation of traditional cross-lingual topic models is their use of language-specific BoW vectors that reside in disjoint feature spaces. Documents from different languages that convey the same semantic content often have non-overlapping vocabularies, forcing models to exert a corrective pressure through integrated loss functions, in a fragile attempt to harmonize their separate topic-word distributions.

\begin{figure}[t]
    \centering
    \includegraphics[width=0.9\linewidth]{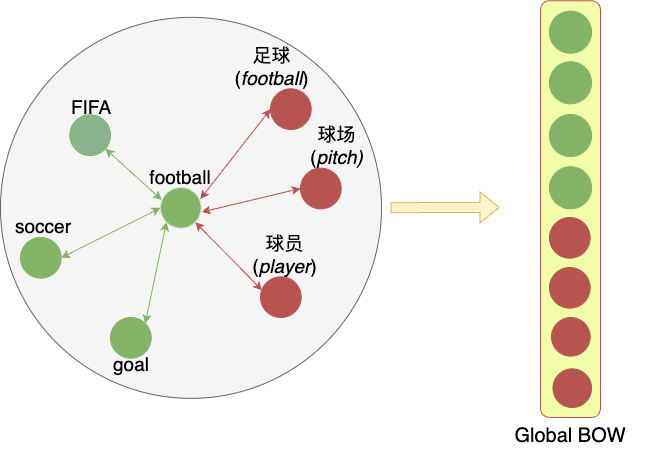}
    \caption{The Polyglot Augmentation mechanism. An original word ("football") is enriched with its nearest intra-lingual (green) and cross-lingual (red) neighbors. This creates a dense Global BOW vector, preemptively injecting cross-lingual information before the encoding step.}
    \label{fig:augment}
\end{figure}

To address this, we introduce \textit{Polyglot Augmentation}, a method that constructs a shared \textbf{Global Context Space} directly at the input level by enriching each document’s Bag-of-Words (BoW) vector with both intra- and cross-lingual neighbors. Given a document-level BoW vector $\mathbf{x}_d^{(l)} \in \mathbb{R}^{|V^{(l)}|}$ and its active word indices $W_d^{(l)} = \{w_i : x_{d,i}^{(l)} > 0\}$, we augment the representation as follows.
First, we retrieve for each active word $w \in W_d^{(l)}$ its top-$k$ most similar intra-lingual neighbors $N_I(w)$ using cosine similarity over word embeddings in language $l$, and cross-lingual neighbors $N_C(w)$ using similarity between embeddings across languages. The enriched BoW vectors are then defined as:
\begin{align*}
\tilde{x}_{d,j}^{(l)} &= \sum_{w \in W_d^{(l)}} \left( [w_j^{(l)} = w] + [w_j^{(l)} \in N_I(w)] \right) \\
\tilde{x}_{d,j}^{(\neg l)} &= \sum_{w \in W_d^{(l)}} [w_j^{(\neg l)} \in N_C(w)]
\end{align*}
where $[\cdot]$ denotes the Iverson bracket, which evaluates to 1 if the condition holds, and 0 otherwise. Finally, the global input vector is the concatenation of the intra- and cross-lingual components in a fixed layout:
\[
\mathbf{g}_d^{(l)} =
\begin{cases}
\left[ \tilde{\mathbf{x}}_d^{(l)} \mid \tilde{\mathbf{x}}_d^{(\neg l)} \right], & \text{if } l = \text{lang}_1 \\
\left[ \tilde{\mathbf{x}}_d^{(\neg l)} \mid \tilde{\mathbf{x}}_d^{(l)} \right], & \text{if } l = \text{lang}_2
\end{cases}
\]
This yields a cross-lingually comparable representation in the joint vocabulary space $\mathbb{R}^{|V^{(1)}| + |V^{(2)}|}$, ensuring that semantically related documents share overlapping features even across languages. This process is illustrated in Figure~\ref{fig:augment}.

This construction yields document vectors that are cross-lingually comparable. For example, documents in $\text{lang}_1$ and $\text{lang}_2$ about the same topic (e.g., ``football'') will include overlapping features such as \text{soccer}, \text{goal}, \text{stadium}, and \text{player}. Consequently, the global encoder no longer needs to infer alignment—it directly observes semantic proximity in a unified input space. This transformation turns alignment from a challenge of bridging separate language spaces into an intrinsic property of the model’s input.

\subsection{Internal Consistency via KL Divergence}
\label{subsec:KL}
Our dual-pathway architecture, with its language-specific \textbf{local encoders} and shared \textbf{global encoder}, risks learning divergent latent spaces that could compromise cross-lingual consistency. To enforce internal alignment, we introduce a regularization term, $\mathcal{L}_{KL}$, that minimizes the Kullback-Leibler (KL) divergence between the local and global posterior distributions for each document:
\[
\mathcal{L}_{KL} = KL(q(z_d^{(l,\text{local})}|x_d^{(l)}) \ || \ q(z_d^{(l,\text{global})}|g_d^{(l)}))
\]
This loss serves a dual purpose. Primarily, it pulls the language-specific representations ($\theta_{\text{local}}$) towards the shared semantic space defined by the global encoder. This ensures the encoders learn meaningful and high-quality topic proportion representations ($\theta$), providing a crucial counterbalance to the strong structural constraints imposed by our unified decoder design.

\subsection{Topic Synchronization via a Unified Decoder}

A core innovation of \textsc{GloCTM} lies in its direct structural enforcement of topic alignment. Instead of applying alignment losses to indirectly connect separate language pathways, the model builds alignment directly into the decoder’s parameterization

\begin{figure}
    \centering
    \includegraphics[width=1\linewidth]{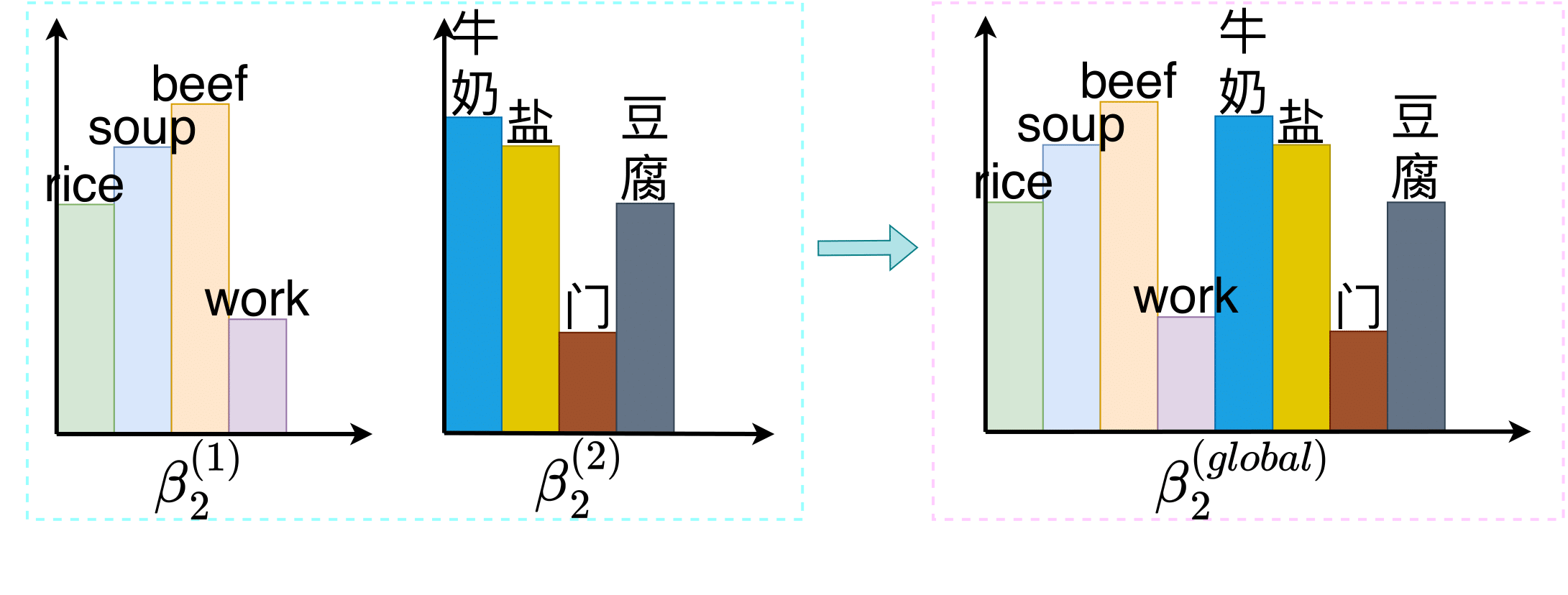}
    \caption{Formation of a unified 'Food' topic (Topic \#2) in GloCTM. The global topic vector is constructed by concatenating the local topic vectors from English ($\beta_2^{(1)}$) and Chinese ($\beta_2^{(2)}$), enforcing semantic alignment by design.
}
    \label{fig:topic_concatenation}
\end{figure}

Specifically, while language-specific topic-word matrices $\beta^{(l)} \in \mathbb{R}^{K \times |V^{(l)}|}$ are used for local views, the global decoder constructs a unified topic-word matrix via horizontal concatenation:
\[
\beta^{(\text{global})} = [\beta^{(1)} \mid \beta^{(2)}]
\]
This design ensures that each topic $k$ is represented by a single, contiguous vector spanning the joint vocabulary, inherently coupling the two languages. An example of this mechanism is illustrated in Figure~\ref{fig:topic_concatenation}. 

The enriched input vector $g_d$ contains words from both vocabularies, but all of them describe the same document and topic. As a result, the global decoder must reconstruct a multilingual signal that consistently represents a single underlying theme. This is achieved using a shared topic vector $\theta_{\text{global}}$ and the unified matrix $\beta^{(\text{global})}$, where each row spans both vocabularies. To minimize reconstruction loss, the decoder is compelled to assign high weights to topically relevant words in \emph{both} languages within the same topic row. If the two halves of a row represent different semantics—e.g., ``Food'' in one language and ``Sports'' in the other—reconstruction will fail, enforcing strong alignment across languages as a structural necessity.
\subsection{Semantic Knowledge Distillation via Centered Kernel Alignment (CKA)}

The rise of multilingual pre-trained language models (MPLMs) provides an opportunity to infuse topic models with deep semantic knowledge. Although our architecture enforces structural alignment, the topic proportion vectors may lack semantic grounding. To enhance their expressiveness, we distill information from MPLMs. However, the mismatch between the \(K\)-dimensional topic space and the \(M\)-dimensional embedding space makes direct similarity comparison ineffective.

We therefore employ \textbf{Centered Kernel Alignment (CKA)} \cite{cortes2012algorithms, kornblith2019similarity}, a robust metric ideal for comparing such disparate representations because it aligns the underlying geometry of the spaces rather than the raw vectors. The distillation objective minimizes the CKA loss between the inferred local topic proportion matrix ($\Theta$) and the MPLM embedding matrix ($E$):
\begin{equation*}
\mathcal{L}_\text{CKA} = 1 - \text{CKA}(\Theta, E)
\label{eq:cka_loss}
\end{equation*}
where CKA is the normalized Hilbert-Schmidt Independence Criterion (HSIC) between the Gram matrices of the two representations ($\mathbf{K}=\Theta\Theta^T, \mathbf{L}=EE^T$):
\begin{equation*}
    \text{CKA}(\Theta, E) = \frac{\text{HSIC}(\Theta, E)}{\sqrt{\text{HSIC}(\Theta, \Theta)\text{HSIC}(E, E)}}
\label{eq:cka_formula}
\end{equation*} 

This process infuses our topics with high-level semantic knowledge, ensuring they are interpretable and meaningfully grounded across languages.
\subsection{Training Objective}

\textbf{VAE Objective per Pathway.} Each pathway in our dual-pathway framework is an instance of a VAE-based topic model. Following the standard Evidence Lower Bound (ELBO) objective \cite{kingma2013vae}, the loss for a single pathway processing an input $\mathbf{x}$ is formulated as:
\begin{equation*}
    \mathcal{L}_{\text{VAE}}(\mathbf{x}) = - \mathbb{E}_{q(\mathbf{z}|\mathbf{x})}[\log p(\mathbf{x}|\mathbf{z})] + \mathrm{KL}(q(\mathbf{z}|\mathbf{x}) || p(\mathbf{z}))
    \label{eq:vae_loss}
\end{equation*}
The first term measures the reconstruction error (negative log-likelihood), while the second term is the KL divergence that regularizes the approximate posterior $q(\mathbf{z}|\mathbf{x})$ to be close to the prior $p(\mathbf{z})$. 

\textbf{Overall Objective for GloCTM.} The final training objective integrates the VAE losses from the global and local pathways, along with two alignment regularizers, $\mathcal{L}_{\text{KL}}$ and $\mathcal{L}_{\text{CKA}}$. Letting $\Phi$ denote all model parameters, the full model is trained end-to-end by minimizing.
\begin{equation*}
\min_{\Phi} \mathcal{L} = \mathcal{L}_{\text{VAE}}^{(\text{global})} + \sum_{l \in \{1,2\}} \mathcal{L}_{\text{VAE}}^{(l, \text{local})} + \lambda_1 \mathcal{L}_{\text{KL}} + \lambda_2 \mathcal{L}_{\text{CKA}}
\end{equation*}

\section{Experiment}
\label{sec:Experiment}

\begin{table*}[!htbp]
    \centering
    \small
    \setlength{\tabcolsep}{.1 mm}
    \begin{tabular}{l*{3}{@{\hspace{0.9mm}}ccc}} 
        \toprule
        \multirow{2}{*}{{Model}} 
        & \multicolumn{3}{c}{{EC News}} 
        & \multicolumn{3}{c}{{Amazon Review}} 
        & \multicolumn{3}{c}{{Rakuten Amazon}} \\
        \cmidrule(lr){2-4} \cmidrule(lr){5-7} \cmidrule(lr){8-10}
        & CNPMI & TU & TQ & CNPMI & TU & TQ & CNPMI & TU & TQ \\
        \midrule
        MCTA     
        & $0.025$ & $0.489$ & $0.012$ 
        & $0.028$ & $0.319$ & $0.009$ 
        & $0.021$ & $0.272$ & $0.006$ \\
        MTAnchor 
        & $-0.013$ & $0.192$ & $0.000$ 
        & $0.028$ & $0.323$ & $0.009$ 
        & $-0.001$ & $0.214$ & $0.000$ \\
        NMTM     
        & $0.028_{\pm0.004}$ & $0.840_{\pm0.016}$ & $0.023_{\pm0.004}$ 
        & $0.043_{\pm0.003}$ & $0.673_{\pm0.028}$ & $0.029_{\pm0.003}$ 
        & $0.010_{\pm0.001}$ & $0.706_{\pm0.022}$ & $0.007_{\pm0.001}$ \\
        InfoCTM  
        & $0.045_{\pm0.002}$ & $0.926_{\pm0.005}$ & $0.041_{\pm0.002}$ 
        & $0.036_{\pm0.004}$ & $0.940_{\pm0.010}$ & $0.034_{\pm0.004}$ 
        & $\underline{0.033_{\pm0.001}}$ & $0.861_{\pm0.008}$ & $\underline{0.028_{\pm0.001}}$ \\
        u-SVD
        & $\underline{0.082_{\pm0.003}}$ & $0.830_{\pm0.006}$ & $\underline{0.068_{\pm0.003}}$ 
        & $\underline{0.055_{\pm0.002}}$ & $0.634_{\pm0.008}$ & $0.035_{\pm0.001}$ 
        & $0.027_{\pm0.001}$ & $0.571_{\pm0.010}$ & $0.015_{\pm0.000}$ \\
        SVD-LR   
        & $\bm{0.083_{\pm0.007}}$ & $0.820_{\pm0.007}$ & $\underline{0.068_{\pm0.006}}$ 
        & $0.053_{\pm0.003}$ & $0.627_{\pm0.013}$ & $0.034_{\pm0.002}$ 
        & $0.026_{\pm0.002}$ & $0.558_{\pm0.010}$ & $0.015_{\pm0.001}$ \\
        XTRA
        & $0.072_{\pm0.002}$ & $\underline{0.982_{\pm0.002}}$ & $\bm{0.070_{\pm0.002}}$
        & $0.052_{\pm0.003}$ & $\bm{0.976_{\pm0.003}}$ & $\underline{0.050_{\pm0.003}}$
        & $0.029_{\pm0.002}$ & $\bm{0.971_{\pm0.007}}$ & $0.027_{\pm0.002}$ \\
        \midrule
        \textbf{GloCTM}
        & $0.071_{\pm0.004}$ & $\bm{0.985_{\pm0.009}}$ & $\bm{0.070_{\pm0.004}}$ 
        & $\bm{0.058_{\pm0.002}}$ & $\underline{0.958_{\pm0.004}}$ & $\bm{0.056_{\pm0.002}}$ 
        & $\bm{0.040_{\pm0.004}}$ & $\underline{0.925_{\pm0.006}}$ & $\bm{0.037_{\pm0.002}}$ \\
        \bottomrule
    \end{tabular}
    \caption{Topic quality scores (CNPMI for coherence, TU for diversity, and TQ = $\max(\text{CNPMI}, 0) \times \text{TU}$ across three datasets. {Bold} and {underline} indicate best and second-best scores, respectively.
    }
    \label{tab_combined_results}
\end{table*}

\subsection{Dataset}
We evaluated our models on three benchmark datasets: EC News \cite{WuLZM20}, containing English–Chinese news articles across six categories; Amazon Review \cite{YuanDY18}, an English–Chinese review dataset reformulated as a binary classification task (five-star = 1, others = 0); and Rakuten Amazon \cite{YuanDY18}, comprising Japanese Rakuten and English Amazon reviews, also cast as a binary rating-based classification task.

\subsection{Evaluation Metrics}
We evaluated topic quality and utility using a multi-faceted approach. Intrinsic quality was measured by cross-lingual coherence (CNPMI)~\cite{cnpmi, npmi} and topic diversity (TU)~\cite{tu}, with both metrics computed using the top 15 words of each topic, along with their resulting Topic Quality (TQ)~\cite{dieng2020etm}. Practical utility was assessed via SVM classification in intra-lingual (-I) and cross-lingual (-C) settings using the topic distributions as features, following prior work~\cite{YuanDY18, infoctm}. Finally, inspired by~\citet{llm_eval}, we used Large Language Model (LLM) ratings for a human-centered evaluation of coherence and alignment. 
\subsection{Baseline Models}
We compare GloCTM with representative baselines: MCTA \cite{DBLP:conf/acl/ShiLBX16} (probabilistic cross-lingual topic modeling), MTAnchor \cite{YuanDY18} (anchor-word alignment), NMTM \cite{WuLZM20} (neural shared-vocabulary alignment), and InfoCTM \cite{infoctm} (mutual-information–based alignment). We also include u-SVD and SVD-LR \cite{refining}, clustering methods on multilingual embeddings, and XTRA \cite{xtra}, a cross-lingual topic model improving coherence and alignment.

\subsection{Overall Topic Quality}
As shown in Table \ref{tab_combined_results}, GloCTM is overall dominant compared to the baselines in terms of topic quality—assessed by coherence (CNPMI), diversity (Topic Uniqueness, TU), and the combined Topic Quality (TQ) score—consistently achieving the highest TU and TQ scores across all datasets. For instance, on the Rakuten Amazon dataset, its CNPMI (0.040) and TQ (0.037) are markedly higher than the best-performing baselines, InfoCTM (CNPMI = 0.033) and XTRA (TQ = 0.027). Moreover, its qualities (TQ) remain top-tier, achieving the highest scores across all three datasets—EC News (0.070, outperforming InfoCTM at 0.041), Amazon Review (0.056, surpassing XTRA at 0.050), and Rakuten Amazon (0.037, exceeding InfoCTM at 0.028). This strong performance across metrics reinforces GloCTM’s lead. Although TU drops slightly, this mainly comes from minor overlaps in key meaningful words—an expected effect of GloCTM’s tighter semantic vocabularies.
\begin{table}[t]
\centering
\setlength{\tabcolsep}{3.4pt} 
\begin{tabular}{lcccccc}
\toprule
& \multicolumn{2}{c}{Topic Quality} & \multicolumn{4}{c}{Classification} \\
\cmidrule(lr){2-3} \cmidrule(lr){4-7}
Model & CNPMI & TU & EN-I & ZH-I & EN-C & ZH-C \\
\midrule
NMTM & 0.045 & 0.643 & 0.788 & 0.721 & 0.592 & 0.575 \\
InfoCTM & 0.036 & 0.934 & 0.788 & 0.725 & 0.672 & 0.601 \\
\midrule
w/o $\mathcal{L}_{\text{KL}}$ & 0.058 & 0.949 & 0.803 & 0.722 & 0.708 & 0.640 \\
w/ $\mathcal{L}_{\text{sim}}$ & \textbf{0.059} & 0.942 & 0.808 & 0.725 & 0.758 & 0.641 \\
w/o $\mathcal{L}_{\text{CKA}}$ & 0.054 & \textbf{0.961} & 0.807 & 0.721 & 0.718 & 0.638 \\
\midrule
\textbf{GloCTM} & 0.058 & 0.958 & \textbf{0.818} & \textbf{0.728} & \textbf{0.763} & \textbf{0.642} \\
\bottomrule
\end{tabular}
\caption{Ablation study of GloCTM on Amazon Review. Scores are means of five seeds; column bests are in {bold}.}
\label{tab:ablation_amazon}
\end{table}

\begin{figure}[t]
\centering
\includegraphics[width=1.\linewidth]{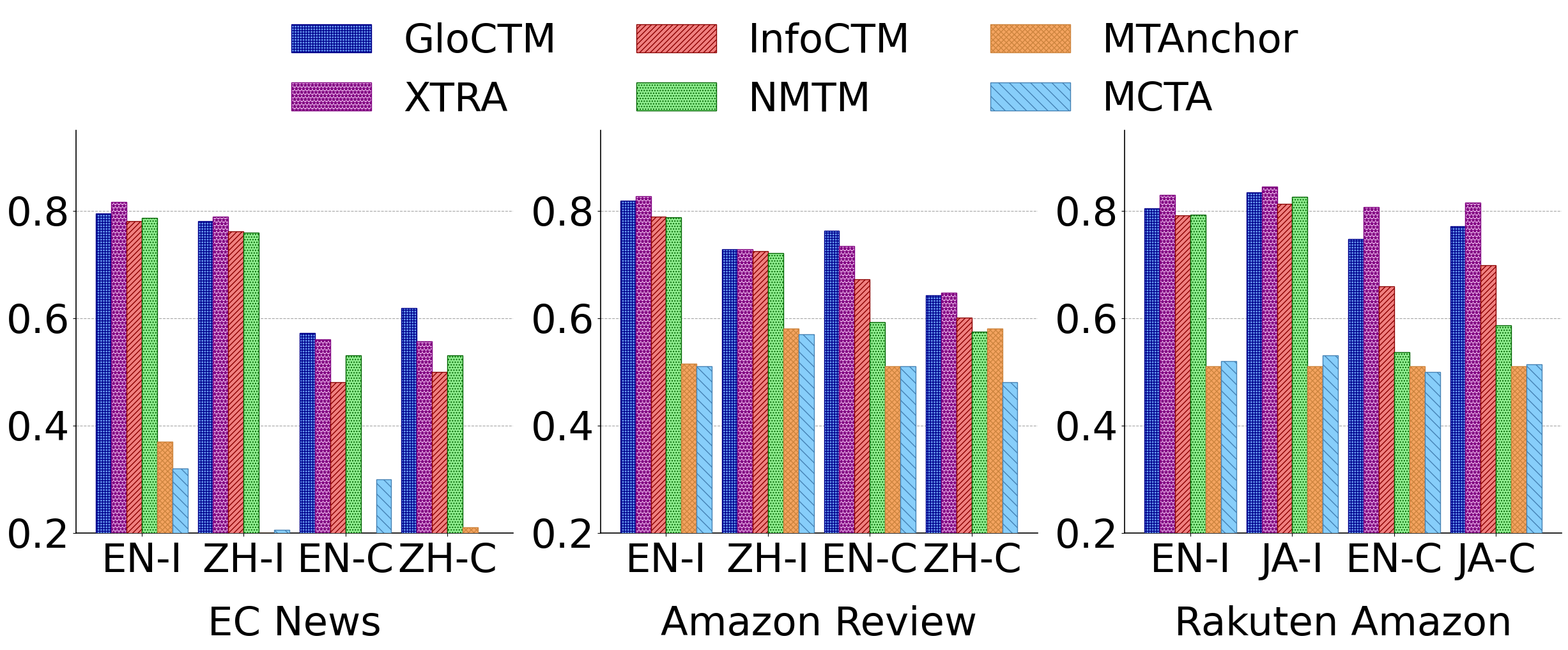}
\caption{Classification performance on the three datasets. Except for MCTA and MTAnchor (results reported in \cite{infoctm}), all scores are averaged over 5 runs.}
\label{fig:all_datasets_cls}
\end{figure}
\begin{figure}[t]
    \centering
    \begin{subfigure}[b]{1\linewidth}
        \centering
        \includegraphics[width=\linewidth]{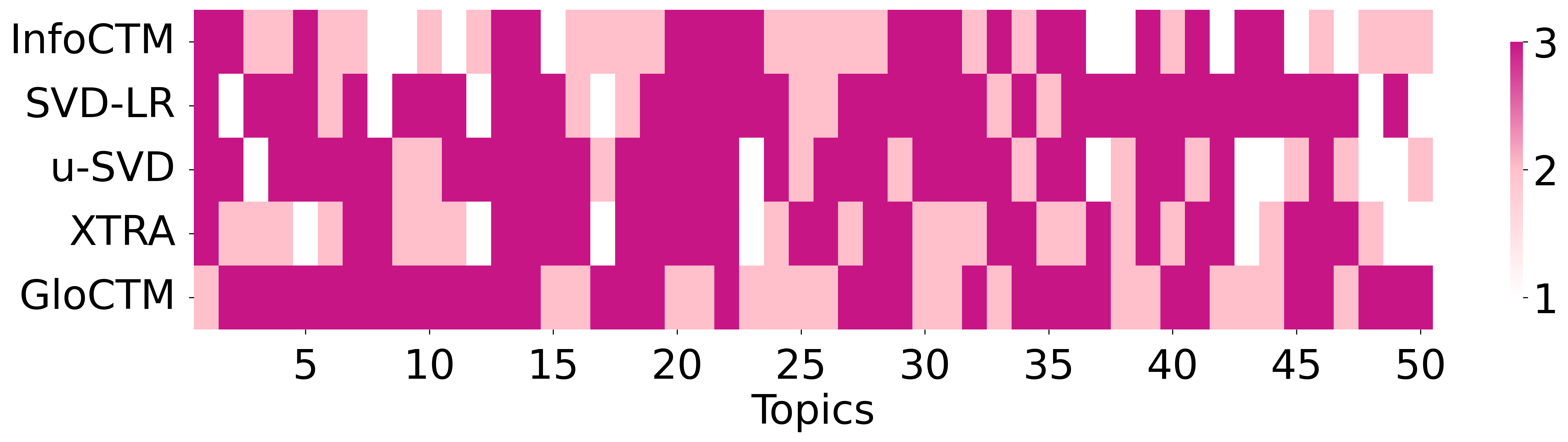}
        \caption{Cross-lingual semantic similarity on EC News dataset.}
        \label{fig:A1}
    \end{subfigure}
    \vspace{1em}

    \begin{subfigure}[b]{1\linewidth}
        \centering
        \includegraphics[width=\linewidth]{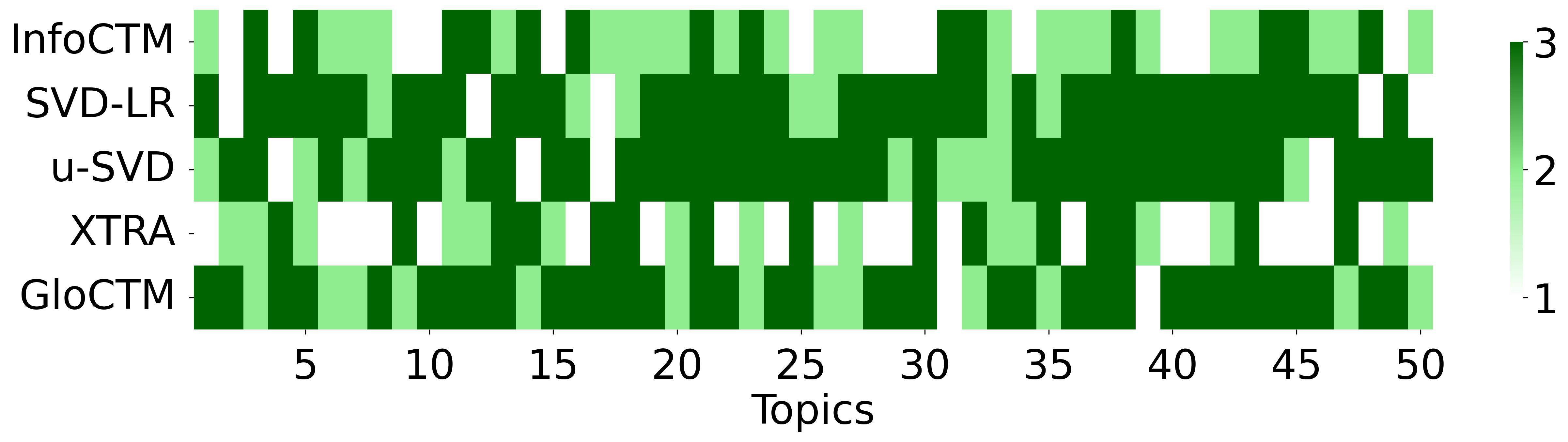}
        \caption{Cross-lingual semantic similarity on Amazon Review dataset.}
        \label{fig:A2}
    \end{subfigure}
    \vspace{1em}

    \begin{subfigure}[b]{1\linewidth}
        \centering
        \includegraphics[width=\linewidth]{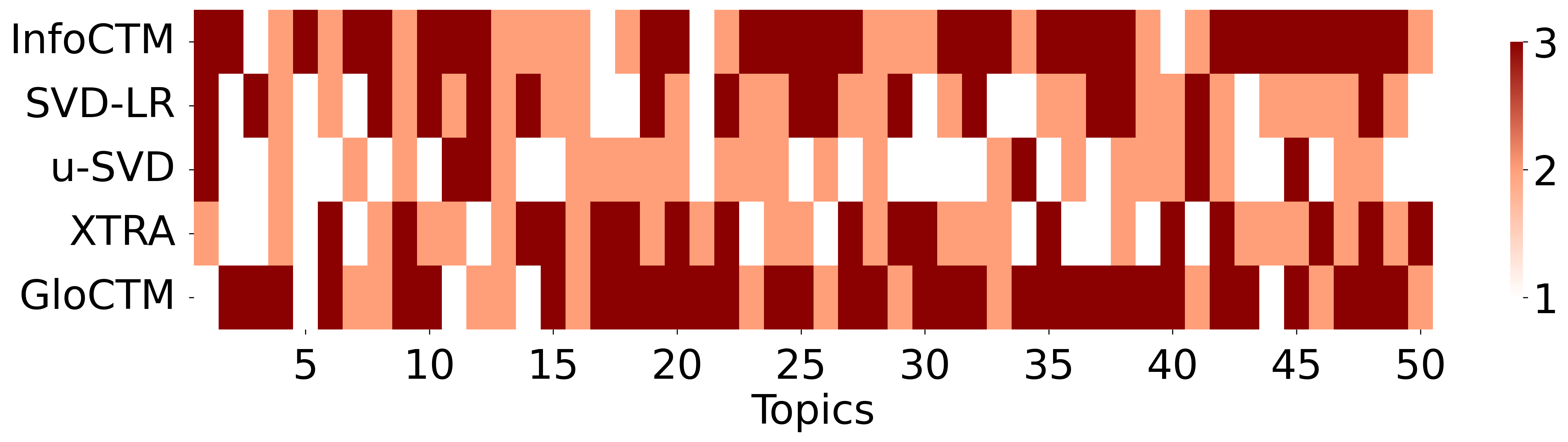}
        \caption{Cross-lingual semantic similarity on Rakuten Amazon dataset.}
        \label{fig:A}
    \end{subfigure}
    
    \caption{LLM-based cross-lingual semantic similarity on Amazon Review, EC News, and Rakuten-Amazon; darker shades indicate higher scores (1–3). Values are means of four runs, rounded.}

    \label{fig:all_llm_evals}
\end{figure}

\begin{table}[t]
    \centering
    \includegraphics[width=.96\linewidth]{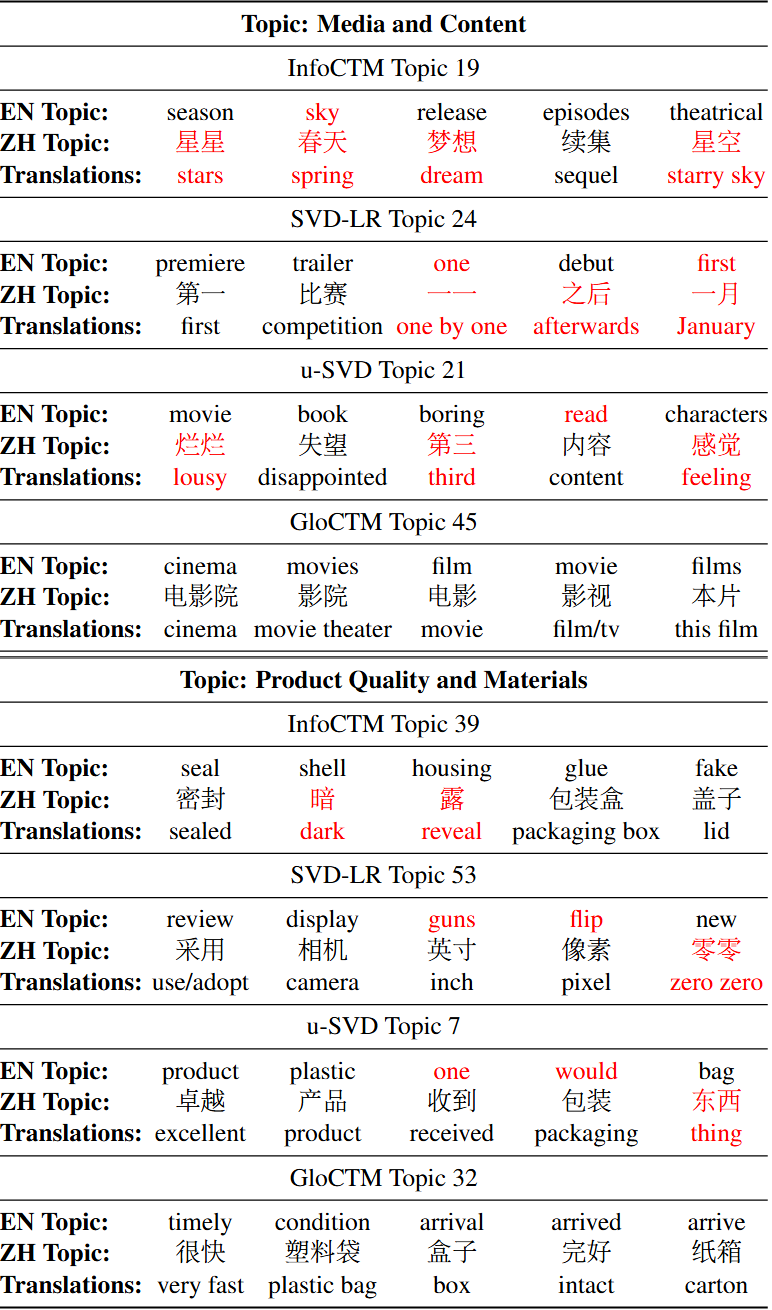}
    \caption{Cross-lingual topic comparison for InfoCTM, SVD-LR, u-SVD, and GloCTM; red words indicate noise or misalignment.}
    \label{fig:topic_comparison}
\end{table}

\label{fig:llm_heatmap}

\subsection{Evaluation of Topic Distributions via Classification}

To evaluate the quality and transferability of document--topic distributions, we use each model's topic vectors \(\theta\) as features for SVM-based text classification, considering both intra-lingual (-I) setups (train/test in the same language) and cross-lingual (-C) setups (train on one language, test on another), as shown in Figure~\ref{fig:all_datasets_cls}. GloCTM remains competitive with classification-focused baselines such as XTRA, showing a slight but consistent advantage over the remaining baselines in intra-lingual settings (EN-I, ZH-I, JA-I), and a clearly superior performance in cross-lingual settings (EN-C, ZH-C, JA-C). Across all three datasets---EC News, Amazon Review, and Rakuten Amazon---GloCTM surpasses every baseline by a substantial margin in cross-lingual transfer.

\subsection{LLM-Assisted Evaluation of Topic Coherence and Alignment}
Following recent work on LLM-based topic evaluation \cite{llm_eval}, we employ large language models to assess cross-lingual semantic similarity between aligned topic pairs, using a 1–3 scoring scale averaged over four independent runs; the final reported score for each topic pair is then rounded to the nearest integer to ensure consistency across evaluations. As shown in Figure~\ref{fig:all_llm_evals}, on the two English–Chinese datasets, GloCTM achieves performance comparable to strong baselines such as SVD-LR and u-SVD while exhibiting greater stability on lower-quality topics, indicating more reliable semantic alignment. The distinction becomes more pronounced on the English–Japanese dataset, where the larger linguistic gap raises alignment difficulty: GloCTM produces a substantially higher number of strong, well-aligned topics and avoids the sharp performance fluctuations observed in competing models. Overall, these LLM-based assessments underscore both the robustness and the cross-lingual reliability of GloCTM’s topic representations.

\subsection{Ablation Study}

To validate our design, we perform an ablation study on the Amazon Review dataset, targeting two key losses: the alignment loss $\mathcal{L}_{\text{KL}}$ and the structural loss $\mathcal{L}_{\text{CKA}}$. We evaluate three variants: (1) removing $\mathcal{L}_{\text{CKA}}$, (2) removing $\mathcal{L}_{\text{KL}}$, and (3) replacing $\mathcal{L}_{\text{KL}}$ with a cosine similarity loss $L_{\text{sim}}$. As shown in Table~\ref{tab:ablation_amazon}, removing $\mathcal{L}_{\text{CKA}}$ degrades classification performance (e.g., EN-C drops to 0.718), confirming its role in latent structure refinement. More critically, removing $\mathcal{L}_{\text{KL}}$ leads to a substantial drop in classification accuracy (EN-C falls to 0.708), highlighting its vital importance in aligning topic distributions. Replacing $\mathcal{L}_{\text{KL}}$ with $\mathcal{L}_{\text{sim}}$ is also observed to be a less effective strategy, as it slightly lowers classification accuracy. These results demonstrate that GloCTM’s effectiveness stems from the synergy between a strong alignment signal ($\mathcal{L}_{\text{KL}}$) and a structure-enforcing signal ($\mathcal{L}_{\text{CKA}}$), leading to the most robust overall performance.


\subsection{Interpreting the Discovered Topics} \label{sec:topic_interpretation}
As shown in Figure~\ref{fig:topic_comparison}, GloCTM generates cleaner, better-aligned topics than InfoCTM, SVD-LR, and u-SVD. In the Media and Content topic, for instance, InfoCTM includes unrelated terms like “sky” and “dream”, SVD-LR suffers from noise and misalignment, and u-SVD mixes in sentiment. In contrast, GloCTM produces focused themes, with English keywords (e.g., “cinema,” “film”) aligned with their Chinese counterparts. Similarly, in the Product Quality topic, GloCTM offers coherent terms (e.g., “timely”, “arrival”), while baselines—especially SVD-LR—introduce off-topic words like “guns” and show topic repetition.

\section{Conclusion}
We propose GloCTM, a dual-pathway framework enforcing intrinsic cross-lingual alignment through Polyglot Augmentation, a shared decoder, and alignment losses (KL, CKA). Experiments show that GloCTM outperforms baselines in topic coherence, diversity, and cross-lingual classification, demonstrating the effectiveness of its alignment mechanism.

\section{Acknowledgements}
Linh Ngo Van is funded by Vietnam National Foundation for Science and Technology Development (NAFOSTED) under grant number 102.05-2025.16. Thien Huu Nguyen has been supported by the NSF grant \# 2239570. He is also supported in part by the Office of the Director of National Intelligence (ODNI), Intelligence Advanced Research Projects Activity (IARPA), via the HIATUS Program contract 2022-22072200003. 

\bibliography{aaai2026}

\end{document}